\documentclass{article}

    \PassOptionsToPackage{numbers, compress}{natbib}

\usepackage[final]{neurips_2021}

\usepackage[utf8]{inputenc} 
\usepackage[T1]{fontenc}    
\usepackage{hyperref}       
\usepackage{url}            
\usepackage{booktabs}       
\usepackage{amsfonts}       
\usepackage{nicefrac}       
\usepackage{microtype}      
\usepackage{xcolor}         
\usepackage{amsmath}
\usepackage{amssymb}
\usepackage{mathtools}
\usepackage{bibentry}
\usepackage{array}
\usepackage{verbatim}
\usepackage{enumitem}
\usepackage{amssymb}
\usepackage{mathtools}
\usepackage{caption}
\usepackage{subcaption}

\usepackage{wrapfig}
\usepackage{makecell}
\usepackage{adjustbox}
\usepackage{colortbl}
\usepackage[ruled,vlined]{algorithm2e}
\SetKwInput{kwInput}{\textbf{Input}}
\SetKwInput{kwInit}{\textbf{Initialization}}

\newcommand{\Tau}{\mathcal{T}}

\newcommand{\calX}[0]{\mathcal{X}}
\newcommand{\calS}[0]{\mathcal{S}}
\newcommand{\calA}[0]{\mathcal{A}}
\newcommand{\calG}[0]{\mathcal{G}}

\title{Consistent Training via Energy-Based GFlowNets for Modeling Discrete Joint Distributions}

\author{Chanakya Ekbote$^1$\thanks{Correspondence to:\texttt{ca10@iitbbs.ac.in}}, Moksh Jain$^1$, Payel Das$^2$, Yoshua Bengio$^{1,3}$\\
$^1$Mila, Universit\'e de Montr\'eal $^2$IBM Thomas J Watson Research Center $^3$ CIFAR AI Chair}

\begin{document}

\maketitle

\begin{abstract}
Generative Flow Networks (GFlowNets) have demonstrated significant performance improvements for generating diverse discrete objects $x$ given a reward function $R(x)$, indicating the utility of the object and trained independently from the GFlowNet by supervised learning to predict a desirable property $y$ given $x$. We hypothesize that this can lead to \textit{incompatibility} between the inductive optimization biases in training $R$ and in training the GFlowNet, potentially leading to worse samples and slow adaptation to changes in the distribution. 
In this work, we build upon recent work on jointly learning energy-based models with GFlowNets and extend it to learn the joint over multiple variables, which we call Joint Energy-Based GFlowNets (JEBGFNs), such as peptide sequences and their antimicrobial activity. Joint learning of the energy-based model, used as a reward for the GFlowNet, can resolve the issues of incompatibility since both the reward function $R$ and the GFlowNet sampler are trained jointly. We find that this joint training or joint energy-based formulation leads to significant improvements in generating anti-microbial peptides. As the training sequences arose out of evolutionary or artificial selection for high antibiotic activity, there is presumably some structure in the distribution of sequences that reveals information about the antibiotic activity. This results in an advantage to modeling their joint generatively vs. pure discriminative modeling. We also evaluate JEBGFN in an active learning setting for discovering anti-microbial peptides.
\end{abstract}

\section{Introduction}



Biological sequence design is a problem of paramount importance for various use cases such as design of new materials, discovering new drugs,  and so on. However, designing these sequences is a time consuming process requiring a lot of domain expertise and many expensive wet lab experiments. Depending on the use case, this can involve searching over an exponentially large combinatorial space (upto orders of $20^{60}$). 

Over the past couple of years, a spate of works \cite{angermueller2019model,Chenthamarakshan2020CogMolTA,moss2020boss,Das2021AcceleratedAD,Wang2021MulticonstraintMG,van2021ampgan} have leveraged recent progress in deep generative methods to aid the search for viable biological sequences. These methods, which fall primarily in the broad areas of reinforcement learning (RL) and Bayesian optimization (BO), typically involve finding sequences that \emph{maximize} certain properties. Consequently, they suffer from one key drawback, the generated sequences have low diversity, i.e. the generated sequences are ``similar". Considering the diverse nature of viable biological targets, and the drug or material discovery process, diversity of generated sequences is a key consideration~\cite{mullis2019diversity,jain2022biological}.

Generative Flow Networks~\citep[GFlowNets;][]{bengio2021flow,bengio2021gflownet} are a family of probabilistic models that alleviate this issue by sampling sequences with probability proportional to the reward, as opposed to reward maximization, resulting in generation of diverse candidates. GFlowNet-AL~\cite{jain2022biological} leveraged GFlowNets in an active learning setting to generate \emph{diverse and novel} biological sequences. In each round of active learning GFlowNet-AL learns a predictor for estimating the reward $R$ from the oracle using the data collected so far. This predictor is then used as the reward for training a GFlowNet. Since the reward function and the GFlowNet are learned separately, we hypothesize that this can lead to \textit{incompatibility} between the reward $R$ and the GFlowNet, potentially leading to poor quality of samples being generated and slow adaptation to changes in the distribution. Moreover, since the generation of sequences is only based on a single reward, this method cannot be used to generate sequences with a specific set of desired properties. For instance, a practical problem of interest is to generate peptide sequences that target a specific class of bacteria, rather than having general toxicity. 

Energy-based models (EBMs)~\cite{lecun2006tutorial,du2019implicit,song2021train} have recently been successfully applied on a wide  variety of probabilistic modelling and inference tasks. In particular, \cite{li2020energy,xie2021active} demonstrate that EBMs adapt quickly to changes in the distribution in continual setting. \cite{zhang2022generative} proposed a framework for jointly training an energy function with a GFlowNet sampler. The GFlowNet samples data proportional to the reward (exp(-energy)) associated with the learned energy function, which models the unnormalised data distribution. The energy function is learnt via contrastive divergence, and the required negative samples are generated by the GFlowNet. Since the reward $R$ (i.e. the energy) and the GFlowNet are both trained via an alternating optimization paradigm, we hypothesize that it reduces the presumed \textit{incompatibility} faced by training them separately.

We extend the framework presented in \cite{zhang2022generative} to model the joint distribution over multiple variables. In particular, we study the problem of learning the joint $p(x,y)$ where $x$ is the data, and $y$ is the corresponding label. Our work offers a GFlowNet-based alternative to prior work on modelling the joint~\cite{pu2018jointgan,chen2019multivariate,che2020your,he2021joint,nie2021controllable,kelly2021directly} for compositional discrete spaces.
To summarize, our key contributions are as follows:

\begin{itemize}[leftmargin=5mm]
    \item We introduce Joint Energy-based GFlowNets (JEB-GFN), an extension of Energy-Based GFlowNets for modelling joint distributions over multiple variables. 
    \item We empirically demonstrate that JEB-GFN can learn the joint distribution over two variables, data $x$ and its corresponding label $y$, and leverage it for conditional generation.
    \item Finally, we evaluate JEB-GFN  for generating peptides with anti-microbial properties in a single round as well as active learning setting.
\end{itemize}

\section{Background}
\label{sec:background}
Generative Flow Networks~\citep[GFlowNets;]{bengio2021flow,bengio2021gflownet} view the problem of sampling discrete objects $x\in \mathcal{X}$ as a problems of learning a control policy. GFlowNets learn a stochastic policy $\pi$ to generate $x$ such that the marginal likelihood of sampling $x \sim \pi(x)$ is proportional to a non-negative reward $R(x)$. In this section we review some basic notation and terminology and refer the reader to \citep{bengio2021gflownet} for a more thorough discussion.

\label{subsec:GFlowNets}
Generative Flow Networks are a probabilistic framework for learning stochastic policies for discrete data. They model the discrete generational process of discrete objects $x \in \calX$, although \citet{bengio2021gflownet} propose a methodology for also generating continuous-valued objects. 
A \textit{compositional} object $x\in \calX$, is modeled as a \emph{trajectory} comprising of a sequence of discrete actions from a set of actions $\calA$. At each step of this trajectory, we obtain a partially constructed object, which we call state $s \in \calS$, which is modified by the next action. The space of possible trajectories is described by a directed acyclic graph (DAG), $\calG = (\calS, \calA)$, whose vertices $\calS$ are the set of all possible states (partially constructed objects along with a special empty state $s_0$) and the edges at each state $\calA$ are actions that modify one state to another. Trajectories starting at $s_0$ and terminating at $x\in \calX$ are called \emph{complete trajectories}, denoted by $\tau \in \Tau$ where $\Tau$ is the set of all possible complete trajectories. Note that while this graph does not need to be instantiated and can be arbitrary in general. We consider the DAG case, so the actions can only be constructive. 

The \emph{trajectory flow} is the unnormalized probability density over the set of all complete trajectories $\tau \in \Tau$; defined as a non negative function $F(\tau)$. For some constant $Z$, we can compute the probability associated with the trajectory as $P_F(\tau) = F(\tau) / Z$. $P_F(s_{t+1} | s_{t})$ is called the forward policy, and sampling iteratively from this forward policy allows one to generate complete trajectories, with a probability proportional to the unnormalized density $F$. We denote the marginal probability of sampling complete trajectories following $P_F$ terminating in $x$ as $\pi(x)$.

The underlying learning problem solved by GFlowNets is to estimate a $P_F$ such that the marginal likelihood of generating object $x$ is proportional to a given non-negative reward $R(x)$. \cite{bengio2021flow} introduced a flow-matching objective to learn this policy. In this work, we focus on the Trajectory Balance view introduced by~\cite{malkin2022trajectory}.
We estimate a forward policy $P_{F}$, a backward policy  $P_{B}$ and a partition function $Z$. The backward policy $P_B$ is a distribution over the set of parents given an input state $s$, which prescribes how to sample trajectories starting at any terminal state $x$ and terminating in $s_0$, i.e. ways in which $x$ \emph{could have been} constructed. The partition function $Z$ denotes the total reward $\sum_{\tau\in\Tau}F(\tau)=\sum_x R(x)$ and is also estimated during training (by $Z_\theta$ below). The parameters $\theta$ parameterizing $P_F(- | s; \theta), P_B(- | s; \theta), Z_\theta$ are estimated by minimizing the following per-trajectory loss:
\begin{equation}
\label{eq:trajectory_balance}
    \mathcal{L}(\tau;\theta) =  \left(\log \frac{Z_{\theta}\prod_{t=0}^{t=n-1}P_{F}(s_{t+1} | s_{t}; \theta)}{R(s_n)\prod_{t=0}^{t=n-1}P_{B}(s_{t} | s_{t + 1}; \theta)}\right)^2.
\end{equation}

\cite{jain2022biological} introduced GFlowNet-AL, an active learning algorithm leveraging GFlowNets for generating \emph{diverse} candidates that maximize some property of the candidate that is expensive to evaluate (oracle). It consists of several rounds, each consisting of generating a batch of candidates that are evaluated with the oracle. To generate each batch of candidates, it first learns a supervised predictor to approximate the oracle and its epistemic uncertainty using a dataset of past evaluations of the oracle. This predictor (and the epistemic uncertainty associated with it) is then converted via an acquisition function into the reward function for training a GFlowNet. As the GFlowNet and reward are trained separately, we have observed that adaptation is slow over rounds (especially if retraining the reward function from scratch) and can result in inconsistencies between the GFlowNet and the reward. This motivates the approach proposed here and described below.

\section{Energy Based Modelling with GFlowNets}


As introduced in section \ref{sec:background}, the training regime of GFlowNets, requires an explicit reward function $R(x)$. Leveraging GFlowNets for generative modelling would require setting the reward function to the unnormalized probability of a particular sample. In practice, one does not have access to the closed-form solution of this unnormalised probability, but only to data sampled from a target distribution. To combat this issue,  \cite{zhang2022generative} introduced the framework of Energy Based GFlowNets (EB-GFNs), where the reward function for the GFlowNet is set using the output from an energy based model (EBM) $\mathcal{E}_{\phi}(x)$ learned simultaneously. The final products obtained include not only be the trained GFlowNet but also the energy based model  $\mathcal{E}_{\phi}(x)$. 


The EBM (associated with the GFlowNet) is trained via the Contrastive Divergence algorithm \cite{carreira2005contrastive}, where the negative samples $x'$ are sampled from a GFlowNet policy ($x' \sim P_{T}(x')$). Note that, $P_{T}(x')$ denotes marginalizing the forward trajectory distribution $P_{F}(\tau)$ over its non-terminating states. With contrastive divergence, one samples $x'$ from a K-step Metropolis Hastings chain, which starts from a true data sample $x \sim p_{data}(x)$. Thanks to the backward policy $P_B$, one can easily perform such MCMC steps with a GFlowNet~\citep{zhang2022generative}. For a perfectly trained GFlowNet, the terminating probability distribution will be equal to the energy distribution. The updates to $\phi$ (the parameter of the energy model) are made proportional to Eq. \ref{eq:cd_ebfgn}:
\begin{equation}
\label{eq:cd_ebfgn}
    \mathbb{E}_{x \sim p_{data}(x)} \nabla_{\phi}\mathcal{E}_{\phi}(x) -   \mathbb{E}_{x' \sim P_{T}(x')}\nabla_{\phi}\mathcal{E}_{\phi}(x').
\end{equation}


We introduce JEB-GFN, which extends the above formulation to model the joint distribution over multiple variables, in particular, data $x$ and corresponding label $y$. The unnormalized joint probability distribution is denoted by $\mathcal{E}_{\phi}(x, y)$. The parameters of the EBM are updated proportional to

\begin{equation}
    \mathbb{E}_{(x, y) \sim p_{data}(x, y)} \nabla_{\phi}\mathcal{E}_{\phi}(x, y) -   \mathbb{E}_{(x', y') \sim P_{T}(x', y')}\nabla_{\phi}\mathcal{E}_{\phi}(x', y')
\end{equation}

Similar to the EB-GFN formulation, the EBM is trained via contrastive divergence, where a K-Step MCMC is used to generate negative samples, given data sampled from the original data distribution. This new generated sample is accepted or rejected depending on the Metropolis-Hastings rule. 

Concretely, first, a K-Step backward trajectory ($\tau$) is sampled with a multiple calls to the backward policy $P_B$, from $(x, y)$ to create an intermediate data sample / state say $(x_{int}, y_{int})$. From this intermediate data sample, a forward trajectory is sampled ($\tau'$) via a forward policy $P_F$, to a terminal state $(x', y')$. Note that the transition probability from state $x$ to state $x'$ is defined as $P_{B}(\tau|x,y)P_{F}(\tau')$, while the reverse transition probability is defined as $P_{B}(\tau'|x',y')P_{F}(\tau)$. The transition of $(x, y)$ to $(x', y')$ is accepted using the Metropolis-Hastings rule, with the probability

\begin{equation}
\label{eq:mh_acceptance}
    A_{\tau, \tau'} ((x, y) \xrightarrow[]{} (x', y')) \overset{\Delta}{=} min [1,  \frac{e^{-\mathcal{E}_{\phi}(x', y')}P_{B}(\tau|x,y)P_{F}(\tau')}{e^{-\mathcal{E}_{\phi}(x, y)}P_{B}(\tau'|x',y')P_{F}(\tau)}].
\end{equation}

Note that the EBM and the GFlowNet are trained via an alternating optimization routine. The procedure for joint training of the GFlowNet and EBM is described in Algorithm \ref{alg:jebgfn_training}.

\begin{algorithm}
\SetAlgoLined
\kwInput{$D=\{x_i, y_i\}, i=1,\dots,N$: Dataset (size $N$) consisting of data: $x_i$ corresponding to labels: $y_i$\\} 
\kwInit{ Initalize GFlowNet's $P_F$, $P_B$ \& $Z$ with parameters $\theta$ and the energy function $\mathcal{E}_{\phi}$ with parameters $\phi$}
\For{$i=1$ to $T$}{
$\bullet$ Sample forward trajectory $\tau \sim P_F(\tau) $\\
$\bullet$ Compute $P_B(\tau)$ given $\tau$, where $\tau$ is generated from the previous step \\
$\bullet$ Update $\theta$ via the gradient step on $\mathcal{L}_{\theta}(\tau)$ (Eq. \ref{eq:trajectory_balance}) with the reward function being $e^{-\mathcal{E}_{\phi}(x, y)}$ \\
$\bullet$ Generate horizon $K$ depending on iteration $i$ \\
$\bullet$ Sample a $K$-step backward trajectory from $P_B(\cdot|\cdot; \theta)$: $\tau$ = ($[x, y]$ = $S_D$ $\rightarrow$ $S_{D-1}$ $\rightarrow$ $...$ $\rightarrow$ $S_{D-K})$ \\
$\bullet$ Sample a $K$-step forward trajectory from $P_F(\cdot|\cdot; \theta)$: $\tau$ = ($S_{D-K}$ $\rightarrow$ $S_{D-K + 1}'$ $\rightarrow$ $...$ $\rightarrow$ $S_{D}'$ $=$ $[x', y']$) \\
$\bullet$ Accept or reject $[x', y']$ via Eq. \ref{eq:mh_acceptance}; set $[x', y']$ = $[x, y]$ if reject. \\
$\bullet$ Update $\phi$ with gradient $\mathcal{E}_{\phi}(x, y) - \mathcal{E}_{\phi}(x', y')$

}
\caption{JEB-GFN joint training framework}
\label{alg:jebgfn_training}
\end{algorithm}


\section{Empirical Results}




\subsection{Synthetic Data}
In this section, we first analyse JEB-GFNs on a set of synthetic tasks adopted from~\cite{zhang2022generative}. To motivate the discussion, we first demonstrate a failure mode of naive conditional generation with GFlowNet. We then demonstrate that JEB-GFN learns the joint $p(x,y)$ and can generate accurate conditional samples $p(x|y)$.

\textbf{Pitfalls of the Naive Conditional Generation Baseline} 

\begin{figure*}[ht]
    \centering
    \begin{subfigure}[t]{0.33\textwidth}
        \centering
        \includegraphics[width=\textwidth]{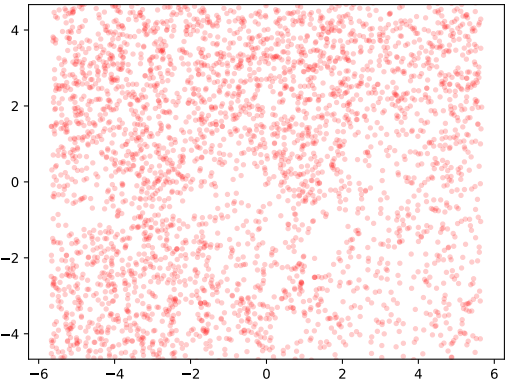}
        \caption{CGFN samples conditioned on label 0}
        \label{subfig:2_moons_mlp_0}
    \end{subfigure}%
    ~ 
    \begin{subfigure}[t]{0.33\textwidth}
        \centering
        \includegraphics[width=\textwidth]{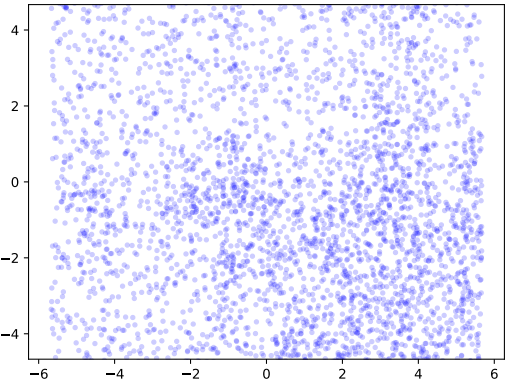}
        \caption{CGFN samples conditioned on label 1}
        \label{subfig:2_moons_mlp_1}
    \end{subfigure}
    ~
    \begin{subfigure}[t]{0.33\textwidth}
        \centering
        \includegraphics[width=\textwidth]{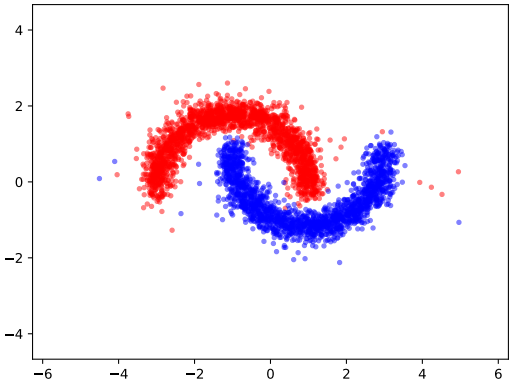}
        \caption{Samples generated by JEB-GFN. Red and blue denote conditioning on labels 0 and 1 respectively}
        \label{subfig:2_moons_ebgfn}
    \end{subfigure}%
    ~ 
    \begin{subfigure}[t]{0.33\textwidth}
        \centering
        \includegraphics[width=\textwidth]{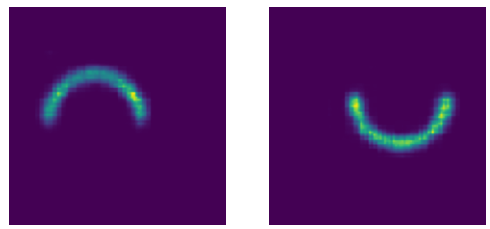}
        \caption{Energy Function learned by JEB-GFN}
        \label{subfig:2_moons_energy}
    \end{subfigure}
    ~
    \caption{(Top) shows a GFlowNet trained with a classifier as reward (CGFN), cannot estimate the underlying distribution. (Bottom) shows that a JEB-GFN can accurately model the underlying data distribution.}
    \label{fig:synthetic_moon}
\end{figure*}

One way to sample from the joint $p(x, y)$ would be to train GFlowNets conditioned on $y$, with the reward being the probability output by a classifier $p(y|x)$, trained on the original dataset $\mathcal{D}$. We evaluate this approach, called CGFN, on the two moons dataset\footnote{\url{https://scikit-learn.org/stable/modules/generated/sklearn.datasets.make_moons.html}}. The procedure is as follows: train a binary classifier on the two moons dataset. Next we train two GFlowNets, one for each class, to generate samples belonging to a particular class. The reward for each of these GFlowNets, was the confidence of the classifier for the generated data for the corresponding class. Following~\cite{zhang2022generative} we convert these 2D floating point values to 16-bit grey-code. The GFlowNets and the Classifier were trained on this 16-bit grey code data. Hence each of these GFlowNets generates a 32 bit sequence autoregressively. At each step the GFN outputs whether a bit at a particular location is either 1 or 0. The samples generated by the GFlowNets are visualized in Figure~\ref{subfig:2_moons_mlp_0} and Figure~\ref{subfig:2_moons_mlp_1}. As one can observe, the samples generated by such an approach are very noisy and do not resemble the original data distribution. A hypothesis to explain this performance is that neural network classifiers tend to be overconfident on out-of-sample data. Consequently, this results in the GFlowNet modelling the "wrong" reward. 

We then trained a JEBGFN on the same setting (i.e. the two moons dataset). Again, these 2D floating point values were converted to 16-bit grey-code. In this case since the JEBGFN outputs both the data, and the class it belongs to, the JEBGFN generates a 33 bit sequence (32 bits for the data and one bit for the class). Again this sequence is generated autoregressively. The samples generated by the GFN and the learned conditional unormalised densities (conditional EBMs) are visualized in Figure~\ref{subfig:2_moons_ebgfn} and Figure~\ref{subfig:2_moons_energy} respectively. Clearly, the JEBGFN is able to learn the original data distribution, given data samples. In sharp contrast to naive approach described above, as JEB-GFN estimates the unnormalised probability density via contrastive divergence, which results in a low density in out-of-sample regions as expected.

\subsubsection{Four Gaussians}
\begin{figure*}[ht]
    \centering
    \begin{subfigure}[t]{0.5\textwidth}
        \centering
        \includegraphics[width=0.75\textwidth]{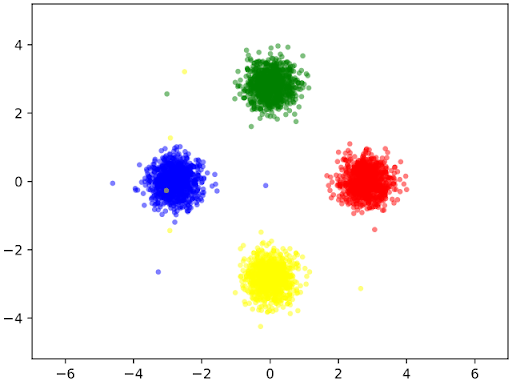}
        \caption{Generated Samples}
        \label{subfig:synthetic_4_gaussians_gaussians}
    \end{subfigure}%
    ~ 
    \begin{subfigure}[t]{0.5\textwidth}
        \centering
        \includegraphics[width=0.75\textwidth]{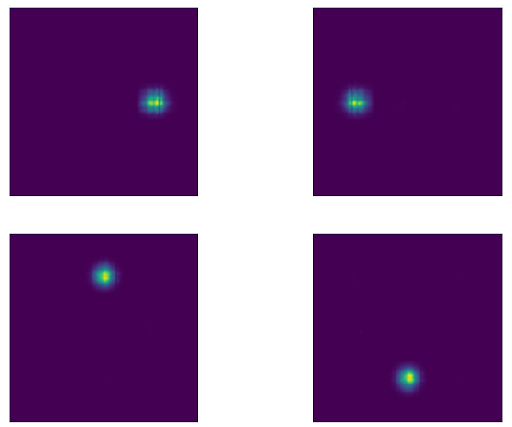}
        \caption{Energy Function learned with JEB-GFN}
        \label{subfig:synthetic_4_gaussians_energy}
    \end{subfigure}
    ~
    \caption{(Left) depicts the samples generated from JEB-GFN, conditioned on each label. Each color represents a different label. (Right) The conditional energy function (conditioned on the label).}
    \label{fig:synthetic_4_gaussians}
\end{figure*}

We now consider a synthetic multi-class task where the data is independently sampled from four 2D Gaussian Distributions, and was assigned a different label depending on the index of the Gaussian that the data was sampled from. The four 2D Gaussians were characterised by the following means: $(2\sqrt{2}, 0), (-2\sqrt{2}, 0), (0, 2\sqrt{2}), (0, -2\sqrt{2})$ and each had a standard deviation of $0.5$. Similar to the previous experiment, the sampled 2D floating point values were converted to 16-bit grey-code. The JEBGFN autoregressively generates a 34 bit sequence (32 bits for the data, 2 bits for the class). The results can be observed in Figure~\ref{fig:synthetic_4_gaussians}. We can observe that the JEBGFN is indeed able to conditionally sample data given a particular class. Moreover, the conditional unnormalised density shows that the underlying distribution is being learned faithfully.

\subsubsection{MNIST}
\begin{figure*}[ht]
    \centering
    \begin{subfigure}[t]{0.5\textwidth}
        \centering
        \includegraphics[width=0.55\textwidth]{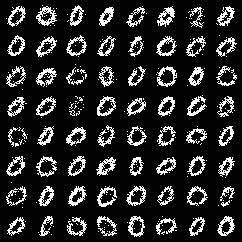}
        \caption{Samples conditioned on label $0$}
        \label{subfig:mnist_0}
    \end{subfigure}%
    ~ 
    \begin{subfigure}[t]{0.5\textwidth}
        \centering
        \includegraphics[width=0.55\textwidth]{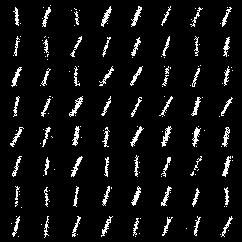}
        \caption{Samples conditioned on label $1$}
        \label{subfig:mnist_1}
    \end{subfigure}
    ~
        \begin{subfigure}[t]{0.5\textwidth}
        \centering
        \includegraphics[width=0.55\textwidth]{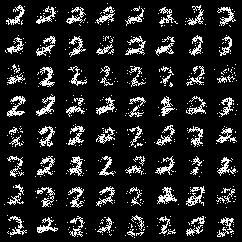}
        \caption{Samples conditioned on label $2$}
        \label{subfig:mnist_2}
    \end{subfigure}%
    ~ 
    \begin{subfigure}[t]{0.5\textwidth}
        \centering
        \includegraphics[width=0.55\textwidth]{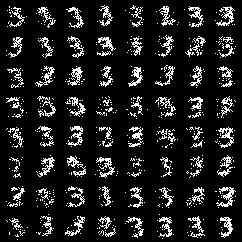}
        \caption{Samples conditioned on label $3$}
        \label{subfig:mnist_3}
    \end{subfigure}
    ~
    \caption{Samples generated with JEB-GFN trained on MNIST}
    \label{fig:synthetic_mnist}
\end{figure*}

The prior two experiments show the efficacy of our method on a dataset where the probability density associated with the dataset is already supplied apriori, i.e. we have the ability to potentially sample infinite datapoints from the data distribution. Two questions that remain unanswered are a) How does our method perform in a limited data setting b) How does our method perform on higher dimensional datasets. Through this experiment we try to answer both these questions. The dataset created for the experiment was obtained by sampling the following digits from the MNIST dataset \cite{deng2012mnist}: $0, 1, 2, 3$. Note that since we don't have the "true" probability density from where the images were sampled from we work only with a limited set of samples (the data points provided in the original dataset). Following~\cite{zhang2022generative} we binarize the original MNIST Data (converting each pixel value to either a 1 or a 0), and then convert it to a 1D array (by concatenating all the rows). Similar to the prior experiments, even in this case the JEB-GFN autoregressively generates a bit sequence, however in this case the length is 786 bits (784 bits for the data and 2 bits for the class)   The results of the samples obtained from the GFlowNet can be observed in Figure~\ref{fig:synthetic_mnist}. One can observe that the JEBGFN is able to generate samples close to the original data distribution. Note that, these results will not be exactly the same as the original dataset because the data has been binarized. 

\subsection{Generating Antimicrobial Peptides}

\subsubsection{Single Round}
\label{subsec:amp_task_single_round}

The goal of this experiment was to generate high scoring peptides (short protein sequences) where the scores are determined based on their anti-microbial properties. We use the dataset from~\cite{jain2022biological}, with 3219 AMPs and 4611 non-AMP sequences obtained from the DBAASP database~\citep{pirtskhalava2021dbaasp}. Note that the sequences considered were of length lower than $50$, and the vocabulary size considered was of size 20 (20 amino acids). The JEBGFN autoregressively generates a tuple $(x,y)$, having a maximum combined length of 51. (50 for $x$ and 1 for the class $y$ which indicates whether a give sequence is antimicrobial). All sequences having a score greater than $0.75$ are assigned the label $1$ corresponding to the positive class, and all the others are assigned the label corresponding to the negative class $0$. The sequences are evaluated by an oracle. Note that, in this case, this is not a bit sequence but a sequence of integers. The range of the integers lie between 0-19, (representing 20 amino acids). Each of these integers are converted to a 256 dimensional learnt embedding, where the position information is also incorporated into this embedding space. We use a 3-layer MLP with 256 units in each hidden layer for both the GFLowNet and EBM. We use the Adam optimizer to train the JEBGFN. We search the learning rate of the GFlowNet in $\{10^{-4}, 10^{-3}, 10^{-2}\}$, the learning rate of the energy function in  $\{10^{-4}, 10^{-3}, 10^{-2}\}$. The coefficient of ${l}_2$ regularization for both the GFlowNet and the EBGFN was searched in $\{0, 0.01, 0.02, 0.05, 0.1\}$.
We compare our method against the following baseline methods: a) GFlowNets~\cite{bengio2021flow} b) GDGDS~\cite{mollaysa2020goal}: a goal-directed conditional generative model , c) SSVAE \cite{kang2018conditional}: a conditional molecule design method, based on a learned semi-supervised VAE. We use the performance (average reward of Top $K$ sequences), diversity (average pairwise distance among the Top $K$ sequences) and novelty (average pairwise distance between Top $K$ generated sequences and the initial dataset) metrics from~\cite{jain2022biological}. 

We can observe that our model performs significantly better than the other baselines across all the metrics. A major point to notice is that our method generates sequences that are more diverse than all the other baselines.  Note that this experiment took $\sim 0.06$ GPU Day(s) to complete, on NVIDIA A100 GPUs.
\begin{table}[h]
\centering
\caption{Single round results on the AMP Task with $K=100$. }
\begin{tabular}{lccc}
\hline
                      & \textbf{Performance} & \textbf{Diversity} & \textbf{Novelty} \\ \hline
\textbf{JEB-GFN}  & $\mathbf{ 0.858 \pm 0.022}$ & $\mathbf{ 33.50 \pm 0.25}$ & $\mathbf{ 24.25 \pm 0.90}$  \\
\textbf{GFlowNet}  & $0.753 \pm 0.023$ & $ 16.43 \pm 4.11$ & $18.38 \pm 3.09$ \\
\textbf{GDGDS}      & $0.712 \pm 0.009 $ & $11.56 \pm 2.71$ & $12.34 \pm 4.12$\\
\textbf{SSVAE}         & $0.743
 \pm 0.012$ & $14.75 \pm 1.98$ & $11.31\pm 1.31$\\ \hline
\end{tabular}
\label{tab:amp_singleround}
\end{table}


\subsubsection{Active Learning}

Through this experiment we try to understand the efficacy of JEB-GFN in the active learning setting, for the AMP Generation task from~\cite{jain2022biological}. We sample $1000$ anti microbial peptides after each round of active learning and evaluate the scores by an oracle. All sequences having a score greater than $0.75$ are assigned the label $1$ corresponding to the positive class, and all the others are assigned the label corresponding to the negative class $0$. Note that this continues for 10 rounds of active learning. After 10 rounds of active learning we pick the Top $K$ sequences (here $K=100$) and then evaluate them. We use the same model details and hyperparameters from Section~\ref{subsec:amp_task_single_round}. We compare our method against the following baseline methods: a) GFlowNet-AL b) DynaPPO ~\cite{angermueller2019model}: A method that uses model-based RL algorithm, for designing biological sequences. c) COMs \cite{trabucco2022design}: a method that relies on generating novel candidates by optimizing known candidates against a learned conservative
model. d) GDGDS e) SSVAE. We use the metrics from~\cite{jain2022biological}, as in Section~\ref{subsec:amp_task_single_round}.
The results for the same can be found in Table~\ref{tab:amp_multi_round}. While the highest score (performance) is obtained by the DynaPPO method, it severly underperforms on the diversity and novelty metrics. Moreoever, though JEB-GFN performs worse than GFlowNet-AL and DynaPPO (in terms of the performance), it does significantly better than all other methods in terms of its diversity ($\sim1.5$x better than its closest baseline). Considering the diverse nature of viable biological targets, we believe that a method producing sequences having a much higher diversity, while also having a high score, would have a higher utility in practice. Note that this experiment took $\sim 1$ GPU Day(s) to complete, on NVIDIA A100 GPUs.


\begin{table}[h]
\centering
\caption{Active Learning Results on the AMP Task with $K=100$. }
\begin{tabular}{lccc}
\hline
                      & \textbf{Performance} & \textbf{Diversity} & \textbf{Novelty} \\ \hline
\textbf{JEB-GFN}  & $0.922 \pm 0.004 $ & $\mathbf{31.83 \pm  0.837}$ & $24.26 \pm  0.965$ \\
\textbf{GFlowNet-AL}  & $\mathbf{0.932 \pm 0.002}$ & $ 22.34 \pm 1.24$ & $\mathbf{ 28.44 \pm 1.32}$ \\
\textbf{DynaPPO}      & $\mathbf{0.938 \pm 0.009}$ & $12.12 \pm 1.71$ & $9.31 \pm 0.69$\\
\textbf{COMs}         & $0.761 \pm 0.009$ & $19.38 \pm 0.14$ & $26.47 \pm 1.3$ \\ 
\textbf{GDGDS}      & $0.785 \pm 0.021$ & $9.41 \pm 1.72$ & $10.73 \pm 2.11$\\
\textbf{SSVAE}       & $0.797 \pm 0.009$ & $10.47 \pm 1.32$ & $10.77\pm 1.39$ \\\hline
\end{tabular}
\label{tab:amp_multi_round}
\end{table}


\section{Conclusion and Future Work}

Motivated from fact that generating viable biological sequences is a time consuming and laborious process, we extend EB-GFN to learn the joint over multiple variables. We validated the efficacy of our method over both synthetic and real-world tasks. Moreover, we have also shown the effectiveness of JEB-GFN in the active learning setting. One limitation of the above method is that the sequences generated can only consist of discrete values. We believe that modelling the joint over both discrete and continuous values would lead to the generation of sequences (conditionally) with an increased level of granularity, which we leave as future work. We do not foresee any negative societal impact from this work but as with all machine learning algorithms it could potentially be used by nefarious agents.


\bibliography{main}
\bibliographystyle{abbrv}
\appendix

\end{document}